\documentclass[sigconf]{acmart}

\usepackage{multirow}
\usepackage{bm}
\usepackage{subcaption} 
\usepackage{enumitem} 
\definecolor{RED}{RGB}{217, 89, 38}
\definecolor{BLUE}{RGB}{72, 165, 214}
\definecolor{GREEN}{RGB}{44, 160, 44}

\AtBeginDocument{%
  \providecommand\BibTeX{{%
    \normalfont B\kern-0.5em{\scshape i\kern-0.25em b}\kern-0.8em\TeX}}}

\copyrightyear{2022} 
\acmYear{2022} 
\setcopyright{rightsretained} 
\acmConference[SIGIR '22]{Proceedings of the 45th International ACM SIGIR Conference on Research and Development in Information Retrieval}{July 11--15, 2022}{Madrid, Spain}
\acmBooktitle{Proceedings of the 45th International ACM SIGIR Conference on Research and Development in Information Retrieval (SIGIR '22), July 11--15, 2022, Madrid, Spain}
\acmISBN{978-1-4503-8732-3/22/07}
\acmDOI{10.1145/3477495.3532052}

\usepackage{etoolbox}
\makeatletter
\patchcmd{\maketitle}{\@copyrightpermission}{
   \begin{minipage}{0.3\columnwidth}
     \href{http://creativecommons.org/licenses/by/4.0/}{\includegraphics[width=0.90\textwidth]{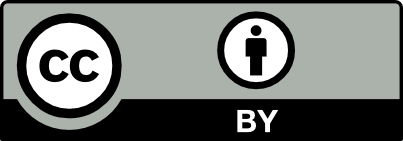}}
   \end{minipage}\hfill
   \begin{minipage}{0.7\columnwidth}
     \href{http://creativecommons.org/licenses/by/4.0/}{This work is licensed under a Creative Commons Attribution International 4.0 License.}
   \end{minipage}
  
   \vspace{5pt}
}{}{}

\makeatother

\begin{document}
\fancyhead{}

\title{Re-thinking Knowledge Graph Completion Evaluation from an Information Retrieval Perspective}

\author{Ying Zhou}
\email{zhouying20@mails.ucas.ac.cn}
\affiliation{%
  \institution{University of Chinese Academy of Sciences}
  \institution{Institute of Software, Chinese Academy of Sciences}
  \city{Beijing}
  \country{China}
}

\author{Xuanang Chen}
\email{chenxuanang19@mails.ucas.ac.cn}
\affiliation{%
  \institution{University of Chinese Academy of Sciences}
  \institution{Institute of Software, Chinese Academy of Sciences}
  \city{Beijing}
  \country{China}
}

\author{Ben He}
\authornote{Corresponding author}
\email{benhe@ucas.ac.cn}
\affiliation{%
  \institution{University of Chinese Academy of Sciences}
  \institution{Institute of Software, Chinese Academy of Sciences}
  \city{Beijing}
  \country{China}
}

\author{Zheng Ye}
\authornotemark[1]
\email{yezheng@scuec.edu.cn}
\affiliation{%
  \institution{South-Central University for Nationalities}
  \city{Wuhan}
  \country{China}
}

\author{Le Sun}
\email{sunle@iscas.ac.cn}
\affiliation{%
  \institution{Institute of Software, Chinese Academy of Sciences}
  \city{Beijing}
  \country{China}
}

\begin{abstract}
Knowledge graph completion (KGC) aims to infer missing knowledge triples based on known facts in a knowledge graph. Current KGC research mostly follows an entity ranking protocol, wherein the effectiveness is measured by the predicted rank of a masked entity in a test triple. The overall performance is then given by a micro(-average) metric over all individual answer entities.
Due to the incomplete nature of the large-scale knowledge bases, such an entity ranking setting is likely affected by unlabelled top-ranked positive examples, raising questions on whether the current evaluation protocol is sufficient to guarantee a fair comparison of KGC systems. 
To this end, this paper presents a systematic study on whether and how the label sparsity affects the current KGC evaluation with the popular micro metrics. Specifically, inspired by the TREC paradigm for large-scale information retrieval (IR) experimentation, we create a relatively ``complete'' judgment set based on a sample from the popular FB15k-237 dataset following the TREC pooling method. 
According to our analysis, it comes as a surprise that switching from the original labels to our ``complete'' labels results in a drastic change of system ranking of a variety of 13 popular KGC models in terms of \textit{micro} metrics.
Further investigation indicates that the IR-like macro(-average) metrics are more stable and discriminative under different settings, meanwhile, less affected by label sparsity.
Thus, for KGC evaluation, we recommend conducting TREC-style pooling to balance between human efforts and label completeness, and reporting also the IR-like \textit{macro} metrics to reflect the ranking nature of the KGC task.
\end{abstract}



\maketitle

\section{Introduction}

Knowledge graphs (KGs) are collections of real-world factual triples, wherein each triple is often presented as a subject $s$, a predicate $p$, and an object $o$.
The rich information leveraged by KGs has become more and more important to various natural language processing (NLP) tasks, such as question answering~\cite{DBLP:conf/naacl/YasunagaRBLL21}, dialogue system~\cite{DBLP:conf/acl/HeBEL17}, or text generation~\cite{DBLP:conf/naacl/Koncel-Kedziorski19}. 
Meanwhile, the inherent incompleteness of KGs has motivated active research to automatically complement the missing links by knowledge graph completion (KGC).
A general KGC approach is to embed entities and relations into a vector space, then assign every possible triple a score to measure the plausibility of its existence.
The \textit{entity ranking} evaluation protocol is usually adopted to assess the performance of KGC models.
Specifically, using the entity ranking protocol, every test triple is converted to two triple questions: \textit{i.e.,} (?, \textit{p, o}) and (\textit{s, p}, ?), where the masked head or tail entity is the prediction target. All candidate entities are then ranked by likelihood of being the masked true entity. Precision of all these masked true entities, e.g., reciprocal rank, are finally averaged to evaluate the performance. Note that a triple question could correspond to more than one answers, e.g., a question (\textit{Trump, visited}, ?) has multiple answers like \textit{China}, \textit{Japan}, and \textit{India}. The evaluation uses answer-wise micro(-average) metrics such that each answer entity contributes equally to the final performance. 

As KGC can be seen as essentially a ranking task with question answering nature~\cite{DBLP:conf/rep4nlp/WangRGBM19}, it shares many similarities with document ranking in information retrieval (IR). However, current KGC evaluation is largely based on extremely sparse labels with micro metrics, unlike in IR evaluation where top-ranked documents are usually densely labeled, and the performance estimation is macro-averaged over individual queries. Such incompatibility has motivated this work to re-think KGC evaluation from an IR perspective.
Actually, as current KGC datasets are all derived from incomplete knowledge bases, there exist inevitably a large quantity of unlabelled positive triples that lead to biased estimation of model performance~\cite{DBLP:conf/rep4nlp/WangRGBM19}. Given the large sizes of existing knowledge bases, it is unfeasible in practice to manually annotate all unobserved positive triples. 
Hence, label sparsity indeed exists in current KGC evaluation, but its impact remains to be investigated.
To bridge this gap, inspired by the TREC pooling method~\cite{voorheesTREC2005}, this paper presents a comprehensive study on label sparsity in KGC evaluation. TREC pooling is a standard evaluation framework typically employed in the TREC experimentation for IR research on large-scale datasets, in which the top-ranked results returned by a diverse set of systems are manually annotated to make up for the label incompleteness. Specifically, this work manually labels a subset of the widely-used FB15k-237 dataset~\cite{toutanova-chen-2015-observed}, coined as \texttt{FB-Test-S-C}.
Our \texttt{FB-Test-S-C} set is composed of 22,492 triples pooled from the top-10 ranked candidates for 1,023 sampled test triples returned by six diverse KGC models, mimicking a relatively complete human judgement set.

After constructing this relatively ``complete'' \texttt{FB-Test-S-C} test set, comprehensive analysis is conducted to answer the following three research questions. {\bfseries RQ1: Does label sparsity indeed affect the KGC evaluation? } According to the re-evaluation results obtained on \texttt{FB-Test-S-C}, drastic changes in system ranking are observed after adding the supplemental labeled triples to the original KGC test set.
The results also raise the concern whether the widely used answer-wise micro metrics are sufficient for the KGC evaluation. Thereby, we introduce the IR-like question-wise macro metrics by treating triple questions as queries, and entity candidates as documents.
With this evaluation setting, another question arises: {\bfseries RQ2: How do the macro metrics differ from the micro metrics?} 
Following recent research on IR evaluation~\cite{DBLP:conf/sigir/Sakai06,DBLP:conf/sigir/AshkanM19,DBLP:conf/ecir/Sakai21}, we comprehensively compare micro and macro metrics in this study.
As shown by the results, the macro metrics are less affected by label sparsity, and are less sensitive to change of test data, while exhibiting better ability in differentiating between different model performances.
Based on all above observations in RQ1 and RQ2, for another question {\bfseries RQ3: What evaluation protocol is suitable for KGC evaluation?}, we suggest to employ a TREC-style pooling for reducing performance biases caused by label sparsity, and additional to the widely-used micro metric, report IR-like macro metrics to reflect the per-question answering performance.
In summary, the main contributions of this paper are three-fold:
\begin{itemize}[leftmargin=*]
    \item A relatively ``complete'' annotated test set based on the widely-used FB15k-237 dataset is constructed following the TREC pooling method. The data and related resources are available online\footnote{ \url{https://github.com/zhouying20/kgc-sparsity}}.
    \item Extensive experiments on our ``complete'' test set show that there is a non-trivial impact of label sparsity on the KGC evaluation. The findings suggest the necessity in conducting TREC-pooling for KGC evaluation in order to negate the impact of label sparsity with acceptable manpower cost.
    \item Comprehensive analyses on the data sensitivity and discriminative power of evaluation metrics indicate the necessity of reporting macro metrics to reflect the per-question performance. 
\end{itemize}

Remainder of this paper is organized as follows. Section~\ref{sec:background} introduces the research background of this study. Section~\ref{sec:construction} describes our construction procedure of the densely annotated \texttt{FB-Test-S-C} dataset using the TREC pooling method. Based on this \texttt{FB-Test-S-C} set, analyses in Section~\ref{sec.sparsity.study} show non-negligible impact of label sparsity on the current KGC evaluation. In Section~\ref{sec.metric.study}, analyses on the stability and discriminative power of evaluation metrics suggest the need for reporting macro metrics. Section~\ref{sec.related.work} recaps the related works, followed by the concluding remarks in Section~\ref{sec.conclusion}. 

\section{Preliminaries} \label{sec:background}

\textbf{Knowledge Graph Completion (KGC).}
Given a set of entities $\mathcal{E}$ and a set of relations $\mathcal{R}$, a Knowledge Graph (KG) is a set of fact triples, namely, $\mathcal{K} = \{(s, p, o)\ |\ s, o\in \mathcal{E}, p\in \mathcal{R}\} \subseteq \mathcal{E} \times \mathcal{R} \times \mathcal{E}$, where $s$, $p$, $o$ stand for subject (entity), predicate (relation) and object (entity), respectively.
The KGC task aims to infer missing triples based on the known facts in the knowledge graph $\mathcal{K}$.
To evaluate a KGC model, the knowledge graph is usually split into training, validation, and test sets, namely, $\mathcal{K} = \mathcal{K}_{train} \cup \mathcal{K}_{valid} \cup \mathcal{K}_{test}$ and $\mathcal{K}_{train} \cap \mathcal{K}_{valid} \cap \mathcal{K}_{test} = \emptyset$.
A general framework for KGC consists of an entity/relation mapper and a scoring function, the former maps the entities and relations into hidden embeddings, and the latter assigns scores to each possible candidate.
Up to now, KGC models can be divided into three categories, namely, translational models (e.g., \cite{DBLP:conf/nips/BordesUGWY13,DBLP:conf/iclr/SunDNT19}), decomposition models (e.g., \cite{DBLP:conf/emnlp/GuoK21,DBLP:conf/icml/NickelTK11}), and neural network models (e.g., \cite{DBLP:conf/aaai/DettmersMS018,DBLP:conf/emnlp/ChenLG0ZJ21}).
Unified frameworks like PyKEEN~\cite{DBLP:journals/corr/abs-2006-13365} and LibKGE~\cite{DBLP:conf/iclr/RuffinelliBG20} are provided to reproduce the previously published results and investigate the impact of hyper-parameters such as loss functions and negative sampling strategies.

\smallskip
\noindent
\textbf{Entity Ranking with Micro Metric.}
KGC models are typically evaluated under the entity ranking (aka. link prediction) protocol, wherein the goal is to find the missing target entity given a \textit{triple question} consisting of a source entity and a relation. KGC models rank all entities in the entity set $\mathcal{E}$ to seek the true object $o$ in $(s, p, ?)$ or the true subject $s$ in $(?, p, o)$. 
To avoid misleading evaluation, a relaxed \textit{filtered} setting \cite{DBLP:conf/nips/BordesUGWY13} is usually adopted, wherein all observed true triples in the graph $\mathcal{K}$
are removed from candidate entities except the one being evaluated, namely, removing all $(s, p, o')$ where $o'\neq o$ and $(s, p, o') \in \mathcal{K}$ when determining the rank of triple $(s, p, o)$.
KGC evaluation usually adopts the \textit{micro} metrics such as the mean rank (\texttt{MR}), mean reciprocal rank (\texttt{MRR}), or \texttt{Hits@K} over individual answers to reflect the performance of KGC models.
Formally, given a true triple $(s, p, o) \in \mathcal{K}_{test}$, all candidate triples $\left\{(s, p, o')~|~o'\in\mathcal{E}~\wedge~(s, p, o') \not\in \mathcal{K} \right\}$ are assigned a score by a KGC model, then $rank(o|s, p)$ is denoted for the \textit{filtered} rank of the true object $o$ among all candidate entities.
If predicting a head entity, $rank(s|p, o)$ is defined similarly.
Thus, micro metrics can be defined uniformly as in Eq.~\ref{eq.micro} to estimate the per-answer performance.
\begin{equation}\label{eq.micro}
\resizebox{0.92\linewidth}{!}{$
    \text{Micro Metric}=\sum_{p\in\mathcal{R}} \left(\frac{c_p}{|\mathcal{K}_{test}|}\cdot\frac{1}{2c_p}\cdot \sum_{(s, p, o)\in\mathcal{K}_{test}} \bigg( f\big(s|p, o\big) + f\big(o|s, p\big) \bigg)\right)
$}
\end{equation}
\noindent wherein $c_p$ is defined as the frequency of occurrences of predicate (relation) $p$ for further analysis on relation distribution in Section~\ref{sec.sparsity.results}; and $f(x) = rank(x)$ for \texttt{MR}; $f(x) = \frac{1}{rank(x)}$ for \texttt{MRR}; $f(x) = 1$ if $rank(x)\leq$ K, else 0 for \texttt{Hits@K}. Hereinafter, we refer to them as \texttt{Micro MR}, \texttt{Micro MRR}, and \texttt{Micro Hits@K}, respectively.

\begin{figure}[t]
\centering
\includegraphics[width=\linewidth, page=1]{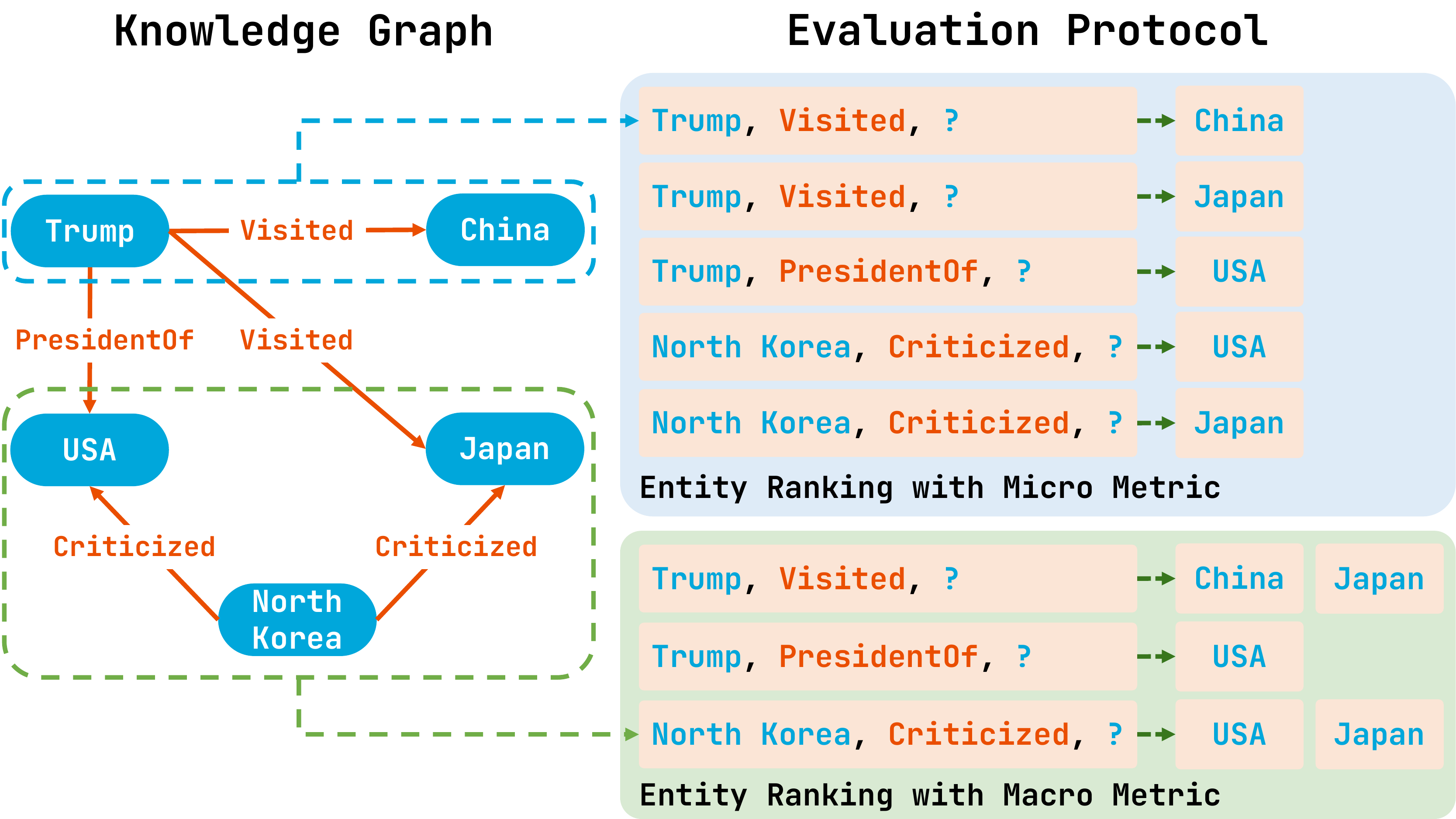}
\caption{Illustration of entity ranking evaluation protocol with micro or macro metrics.
Left side: a small knowledge graph with facts about Donald Trump and the USA.
\textcolor{BLUE}{Blue box}: commonly used entity ranking with micro metric that constructs questions from KG triples for answer-wise evaluations.
\textcolor{GREEN}{Green box}: entity ranking with macro metric that aggregates the same triple questions for IR-like question-wise evaluations.}
\label{fig:triplet_retrieval}
\end{figure}

\smallskip
\noindent \textbf{Entity Ranking with Macro Metric.}\label{sec.entity.marco}
Additional to entity ranking with micro metric, 
we also introduce the common information retrieval evaluation methodology and macro metrics to entity ranking. 
The illustration of entity ranking with micro and macro metrics is shown in Figure~\ref{fig:triplet_retrieval}.
Analogously, in the entity ranking protocol with macro metric, the triple questions (same triple questions are aggregated) play the role of queries, and all target entities that answer the question act as the relevant documents.
Specifically, with regard to a KG, all triple questions in the test set are aggregated and collected as a triple question set $\mathcal{Q}_{test}$.
For example, the triple question \textit{(Trump, visited, ?)} is considered twice in the micro metrics, but they are aggregated and only considered once in the macro metrics.
Then, for each aggregated question $q\in \mathcal{Q}_{test}$, the corresponding target entities in the test set are considered relevant to form the answer set $\mathcal{A}_{q}$, while other entities are considered as unjudged (their relevance is unknown).
When evaluating a KGC model, for each $q\in \mathcal{Q}_{test}$, we rank all entities that have not been seen in the training and validation sets.
Then, we use IR evaluation metrics as implemented in the standard \texttt{trec\_eval}\footnote{\url{https://trec.nist.gov/trec\_eval/}} toolkit that is widely-used for the TREC evaluation~\cite{voorheesTREC2005}, including macro-average reciproal rank (\texttt{MRR}), \texttt{MAP@20}~\cite{DBLP:books/daglib/0021593}, and \texttt{nDCG@20} \cite{DBLP:journals/tois/JarvelinK02}.
To be consistent with micro metrics, we also implement the widely-used \texttt{Hits@K} in a macro setting, wherein, for an aggregated triple question, we regard any answer ranked in top-K candidates as a hit.
Similar to micro metrics in Eq.~\ref{eq.micro}, we also define macro metrics uniformly as in Eq.~\ref{eq.macro}.
\begin{equation}\label{eq.macro}
    \text{Macro Metric}=\frac{1}{|\mathcal{Q}_{test}|}\sum_{q\in\mathcal{Q}_{test}} f(q)
\end{equation}
\noindent wherein $f(q)=1$ if $min(rank(a|q))\leq K,a\in\mathcal{A}_q$, else $f(q)=0$ for \texttt{Hits@K}; $f(q)=\frac{1}{min(rank(a|q))}, a\in\mathcal{A}_q$ for \texttt{MRR}.
To distinguish from the micro metrics, we refer to them as \texttt{Macro MRR}, \texttt{Macro Hits@K}.
As for \texttt{MAP@20} and \texttt{nDCG@20}, please refer to \texttt{trec\_eval} for implementation details.
To be consistent with previous works, in the following experiments, we employ \textit{filtered} setting in both micro and macro evaluations, namely, for macro metric, we only rank candidates that are not observed in the training and validation sets; and for micro metric, we additionally remove other answer entities such that only the target answer entity is reserved.

\begin{table}[tbp]
\caption{Statistics of the original FB15k-237 test set (named  \texttt{FB-Test-O}), the sampled subset (\texttt{FB-Test-S}), and the relatively ``complete'' test set (\texttt{FB-Test-S-C}). The original set does not have any negative triple. Our annotation finds more positive and many negative triples.}
\centering         
\begin{tabular}{l|*{3}{c}}
\toprule
\textbf{Test Set} & \textbf{\#Triples} & \textbf{\#Positives} & \textbf{\#Negatives} \\
\midrule
\texttt{FB-Test-O} & 20,466 & 20,466 & 0 \\
\texttt{FB-Test-S} & 1,023 & 1,023 & 0 \\
\texttt{FB-Test-S-C} & 22,492 & 2,738 & 19,754 \\
\bottomrule
\end{tabular}
\label{tab:testset_statistics}
\end{table}

\section{Data Construction} \label{sec:construction}
As mentioned, there is a sparsity issue in the current KGC datasets. Herein, in order to investigate the influence of label sparsity on the evaluation and comparison of various KGC systems, our work constructs a relatively ``complete'' test set by adding manually labeled triples into one existing ``incomplete'' test set from widely-used FB15k-237 dataset~\cite{toutanova-chen-2015-observed}.
In this section, we first introduce the construction procedure of our ``complete'' test set, followed by data statistics and analyses on the created dataset.

\begin{table*}[tbh]
\caption{Examples of our annotations. The original answers exist in the original FB15k-237 test set. Our annotations are carried out on both positive and \underline{negative} entities.}
\centering         
\resizebox{0.99\linewidth}{!} {%
\begin{tabular}{*{3}{l}}
\toprule
{\bfseries Triple Question} & {\bfseries Original Answers} & {\bfseries Annotations} \\
\midrule
{\itshape (Shia LaBeouf, /people/person/profession, ?)} & {Actor} & Voice Actor, Comedian, \underline{Model}, \underline{Television Director} \\
\midrule
{\itshape (European American, /people/ethnicity/languages\_spoken, ?)} & {American English} & Spanish Language, Danish Language, \underline{Mandarin Chinese} \\
\midrule
{\itshape (United States of America, /location/location/contains, ?)} 
& {Valdosta, Concord} &  Manhattan, Boston, Chicago, \underline{Chelsea}, \underline{Marrakesh} \\
\bottomrule
\end{tabular}
}
\label{tab:annotation_examples}
\end{table*}

\subsection{Annotation Procedure}
We introduce the annotation methodology based on the TREC pooling method in this section. Recent works~\cite{DBLP:conf/emnlp/ChenLG0ZJ21,DBLP:conf/aaai/DettmersMS018,DBLP:conf/iclr/RuffinelliBG20,DBLP:conf/iclr/VashishthSNT20,DBLP:conf/aaai/ZhangCZW20} on entity ranking for KGC almost all adopt two datasets for the evaluation, namely, FB15k-237~\cite{toutanova-chen-2015-observed} derived from Freebase~\cite{DBLP:conf/sigmod/BollackerEPST08}, and WN18RR~\cite{DBLP:conf/aaai/DettmersMS018} derived from WordNet~\cite{fellbaumWordNet1998}.
Between those two datasets, FB15k-237~\cite{toutanova-chen-2015-observed} is larger, which comes with 20,466 test triples (denotes as \texttt{FB-Test-O}), covering a wide range of knowledge extracted from Freebase. We therefore base our annotation on this dataset. As supplementing judgement on this entire test set requires exhaustive manpower, we rather annotate based on a small but representative subset of \texttt{FB-Test-O}.
Specifically, we randomly sample 5\% test triples, which is 1,023 out of 20,466, to obtain a relatively smaller initial triple seed set, denoted as \texttt{FB-Test-S}. Meanwhile, the original relation distribution (actually only 0.0889 KL divergence between the sampled and original relation distribution) is maintained to avoid introducing data bias to this subset, which will be further analyzed in Section~\ref{sec.analysis.complete}.

\smallskip
\noindent
{\bfseries Pooling.}
For every seed triple in the \texttt{FB-Test-S} set, we collect its top-10 predicted triples from a series of KGC models to form the pool of triples waiting for judgement, namely, the top-10 ranked entities for both the head questions $(?, p, o)$ and the tail questions $(s, p, ?)$.
To reduce the bias towards a specific KGC model, we select six diverse baseline systems with regard to their model architectures, including the translational models TransE~\cite{DBLP:conf/nips/BordesUGWY13} and RotatE~\cite{DBLP:conf/iclr/SunDNT19}, the decomposition models ComplEx~\cite{DBLP:conf/icml/TrouillonWRGB16}, DistMult~\cite{DBLP:journals/corr/YangYHGD14a} and RESCAL~\cite{DBLP:conf/icml/NickelTK11}, and the neural model ConvE~\cite{DBLP:conf/aaai/DettmersMS018}, to obtain the less overlapped pooled triples. We use the well-tuned model checkpoints from LibKGE~\cite{DBLP:conf/iclr/RuffinelliBG20}.
Finally, we obtain a pool of triples, consisting of 7,087 entities, 151 relations, and 21,469 triples.

\begin{figure}[t]
\centering
\begin{subfigure}{.95\linewidth}
\includegraphics[width=1.0\linewidth, page=1]{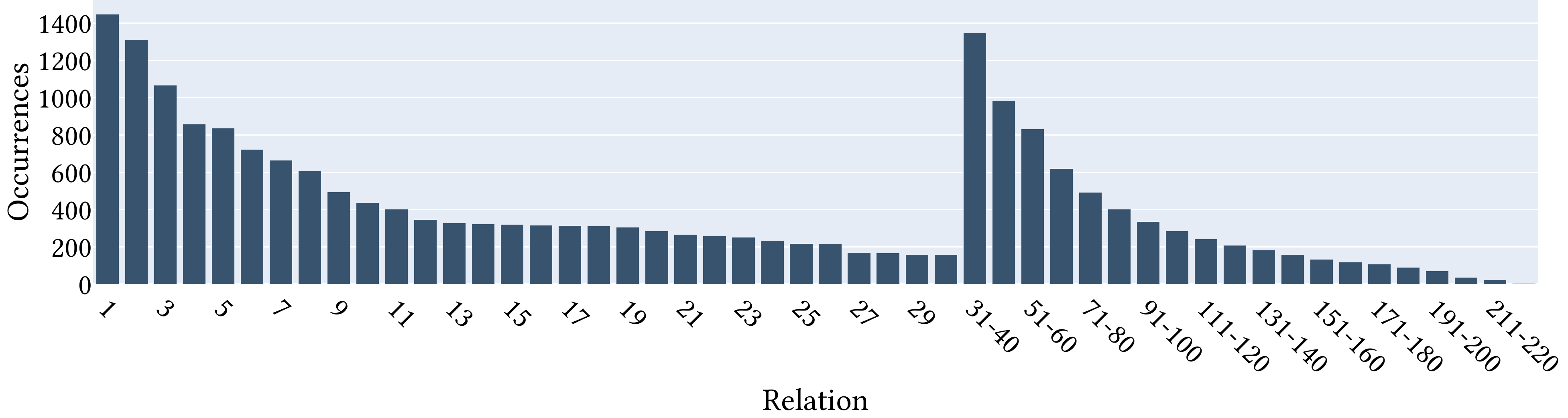}
\caption{\texttt{FB-Test-O}}
\label{fig:sub_dist_origin}
\end{subfigure}

\begin{subfigure}{.95\linewidth}
\includegraphics[width=1.0\linewidth, page=2]{figures/relation_distribution.pdf}
\caption{\texttt{FB-Test-S}}
\label{fig:sub_dist_sample}
\end{subfigure}

\begin{subfigure}{.95\linewidth}
\includegraphics[width=1.0\linewidth, page=3]{figures/relation_distribution.pdf}
\caption{\texttt{FB-Test-S-C}}
\label{fig:sub_dist_complete}
\end{subfigure}
\caption{Relation distribution in the original FB15k-237 test set (\texttt{FB-Test-O}), sampled test subset (\texttt{FB-Test-S}), and human-labeled relatively complete test subset (\texttt{FB-Test-S-C}).}
\label{fig:relation_distribution}
\end{figure}

\smallskip
\noindent
{\bfseries Filtering Trivial Triples.}
To reduce the cost of human annotation, a certain number of trivial triples are automatically filtered, which are largely false triples: 1) the types of the subject and the object are not consistent with the relation; 2) the triples collected from head questions $(?, p, o)$ with obvious one-to-one and one-to-many relations; 3) the triples collected from tail questions $(s, p, ?)$ with obvious one-to-one and many-to-one relations.
Note that we treat all these selected trivial triples as negatives without the need for human annotation.
Thereby, the final pool for later manually labeling contains 12,599 triples.

\smallskip
\noindent
{\bfseries Annotating Triple Questions.}
Akin to~\cite{DBLP:conf/akbc/PezeshkpourT020,DBLP:conf/emnlp/SafaviK20}, we annotate both positive and negative triples in the pool.
In order to facilitate the judgment of the annotator, we first manually create templates to transform triples to natural language questions, e.g., converting (\textit{Donald Trump, /people/person/nationality, USA}) to ``is Donald Trump's nationality USA?''.
During the annotation, annotators were asked to search for information on the Internet and answer Yes or No to each triple question.
In particular, for the reliability of annotations, annotators were asked to provide the source link which contains the information for answering the question. Meanwhile, the provided source link should be in the list of verified data sources including Wikipedia and other domain-specific sources like AllMusic, IMDB, etc.
A total of five graduate students majored in computer science participated in the annotation. Each triple was labeled by two independent annotators. When there is a conflict, a third annotator steps in to provide the final annotation.

Finally, all pooled triples are annotated as positive (Yes) or negative (No), and this set of triples could be considered as a relatively ``complete'' test set, denoted as \texttt{FB-Test-S-C}.
The statistics of these three test sets, namely, \texttt{FB-Test-O}, \texttt{FB-Test-S}, and \texttt{FB-Test-S-C} are summarized in Table~\ref{tab:testset_statistics}.
With adding supplemental annotated 21,469 triples into \texttt{FB-Test-S} set, the relatively ``complete'' \texttt{FB-Test-S-C} set contains much more judged triples, namely, 22,492 triples with 2,738 positives and 19,754 negatives. 
In other words, our annotations mainly complement the number of positive triples (which is directly related to the data sparsity problem), and also a large number of judged negative triples.

\subsection{Analysis of Annotation}\label{sec.analysis.complete}
{\bfseries Annotated Examples.}
To ensure the quality of our annotations, we randomly selected 200 triplets from the newly annotated test set (half positives and half negatives) and gathered the relabeled results from an expert in the KG field. We found that 92\% of the annotations are consistent with the pooled labels.
Additionally, we present three selected annotated examples in Table~\ref{tab:annotation_examples}. As seen in the table, the examples demonstrate that the label sparsity indeed exists in the current FB15k-237 dataset.
Take the triple question {\itshape (European American, /people/ethnicity/languages\_spoken, ?)} for example, if two KGC models assign the same rank for the true answer `{\itshape American English}', while one ranks `{\itshape Mandarin Chinese}' higher and the other ranks `{\itshape Spanish Language}' higher,
according to the current entity ranking protocol, since both `{\itshape Mandarin Chinese}' and `{\itshape Spanish Language}' are unjudged (considered as negative), these two KGC models are given the same performance metric.
However, intuitively, the KGC model that ranks the false negative `{\itshape Spanish Language}' higher should be considered better on this triple question.
Thereby, the evaluation on the test set with incomplete judgement could lead to a misleading comparison of KGC models.

\smallskip
\noindent
{\bfseries Relation Distribution.}
As mentioned, the relation distribution in our sampled \texttt{FB-Test-S} set is maintained as that in the original \texttt{FB-Test-O} test set by random sampling. 
Herein, there are totally 224 different relations in the \texttt{FB-Test-O} set, with different frequencies. Besides, to better display in Figure~\ref{fig:relation_distribution}, low frequency relations are merged, resulting in 50 relation classes.
As shown in Figures~\ref{fig:sub_dist_origin}~and~\ref{fig:sub_dist_sample}, the frequency of all relations in the \texttt{FB-Test-S} set are 
highly analogous to the original \texttt{FB-Test-O} set. 
Specifically, the KL divergence (KLD) between \texttt{FB-Test-S} and \texttt{FB-Test-O} is 0.0889, while the KLD between \texttt{FB-Test-O} and the full FB15k-237 set (including train, validation and test sets, whose relation distribution is not plotted for space reason) is 0.2789. Therefore, the relation distribution in the sampled \texttt{FB-Test-S} set is close enough to the original \texttt{FB-Test-O} set.
Furthermore, we present the relation distribution of our labeled ``complete'' \texttt{FB-Test-S-C} set in Figure~\ref{fig:sub_dist_complete}.
We can see that the overall distribution of \texttt{FB-Test-S-C} has slightly changed, resulting in a relatively larger KLD of 0.2386 to the \texttt{FB-Test-O} set. 
The influence of this changed relation distribution will be investigated in Section~\ref{sec.sparsity.results}.
Note that we can only add new positive entities to one-to-many relations, but not to one-to-one relations, which will also be further discussed in Section~\ref{sec.sparsity.results}.

\section{Study on Label Sparsity}\label{sec.sparsity.study}
In this section, to answer RQ1, we conduct experiments to investigate how the label sparsity affects the KGC evaluation. 
We first introduce the experimental settings in Section~\ref{sec.sparsity.design}, followed by the results and analyses in Section \ref{sec.sparsity.results}, in which inconsistent system rankings are observed with the increase of label completeness.

\subsection{Experimental Design}\label{sec.sparsity.design}
Akin to studies on label sparsity in IR evaluation~\cite{DBLP:conf/sigir/HeMO08,DBLP:conf/sigir/ButtcherCYS07,DBLP:conf/sigir/BuckleyV04,DBLP:conf/sigir/YilmazKA08,DBLP:conf/www/Kirnap0BECY21,DBLP:conf/ictir/RahmanKEL20}, we explore the influence of label sparsity on KGC evaluation by comparing the rankings of KGC systems on an ``incomplete'' test set and its relatively ``complete'' versions.

\smallskip
\noindent
\textbf{Datasets.}
As described in Section~\ref{sec:construction}, we have constructed a relatively ``complete'' \texttt{FB-Test-S-C} test set based on the \texttt{FB-Test-S} set using the pooling technique.
Herein, we carry out experiments on these two paralleled \texttt{FB-Test-S} and \texttt{FB-Test-S-C} test sets, to see the changes in the rankings of KGC systems.
When constructing the \texttt{FB-Test-S-C} set, we observe that the amount of unobserved true triples actually consistently grows with the pooling depth.
Thereby, additional to the full \texttt{FB-Test-S-C} test set, we also investigate the influence of the different levels of judgement completeness to the KGC evaluation, by controlling the pooling depth.

\smallskip
\noindent
\textbf{Metrics.}
As described in Section~\ref{sec:background}, in the entity ranking evaluation, previous works usually use the micro metrics to report the per-answer performance of KGC models.
Thereby, we mainly focus on micro metrics in this part to examine how label sparsity affects comparisons of KGC models.
Herein, five commonly reported micro metrics, including \texttt{Micro Hits@1}, \texttt{Micro Hits@3}, \texttt{Micro Hits@10}, \texttt{Micro MR}, and \texttt{Micro MRR}, are used to evaluate and rank a series of KGC models.
Meanwhile, to answer RQ3, comparisons between micro and macro metrics need to be carried out, so we also provide the results on macro metrics, including \texttt{Macro Hits@10} and \texttt{Macro MRR}.
As for the indicator of the changes in ranking of KGC systems, we choose Kendall's $\tau$~\cite{kendallTau1938} following previous works on IR evaluation~\cite{DBLP:conf/sigir/Sakai06,DBLP:conf/sigir/HeMO08,DBLP:conf/sigir/ButtcherCYS07,DBLP:conf/sigir/BuckleyV04,DBLP:conf/www/Kirnap0BECY21,DBLP:conf/acl/Sakai20}. 
Kendall's $\tau$ is a statistical metric used to verify ordinal association between two measured quantities, ranging from $-1$ to $+1$, wherein $+1$ for totally matching and $-1$ for the opposite.
In other words, smaller Kendall's $\tau$ indicates larger divergence in system rankings.

\begin{figure}[t]
\centering
\includegraphics[width=\linewidth, page=1]{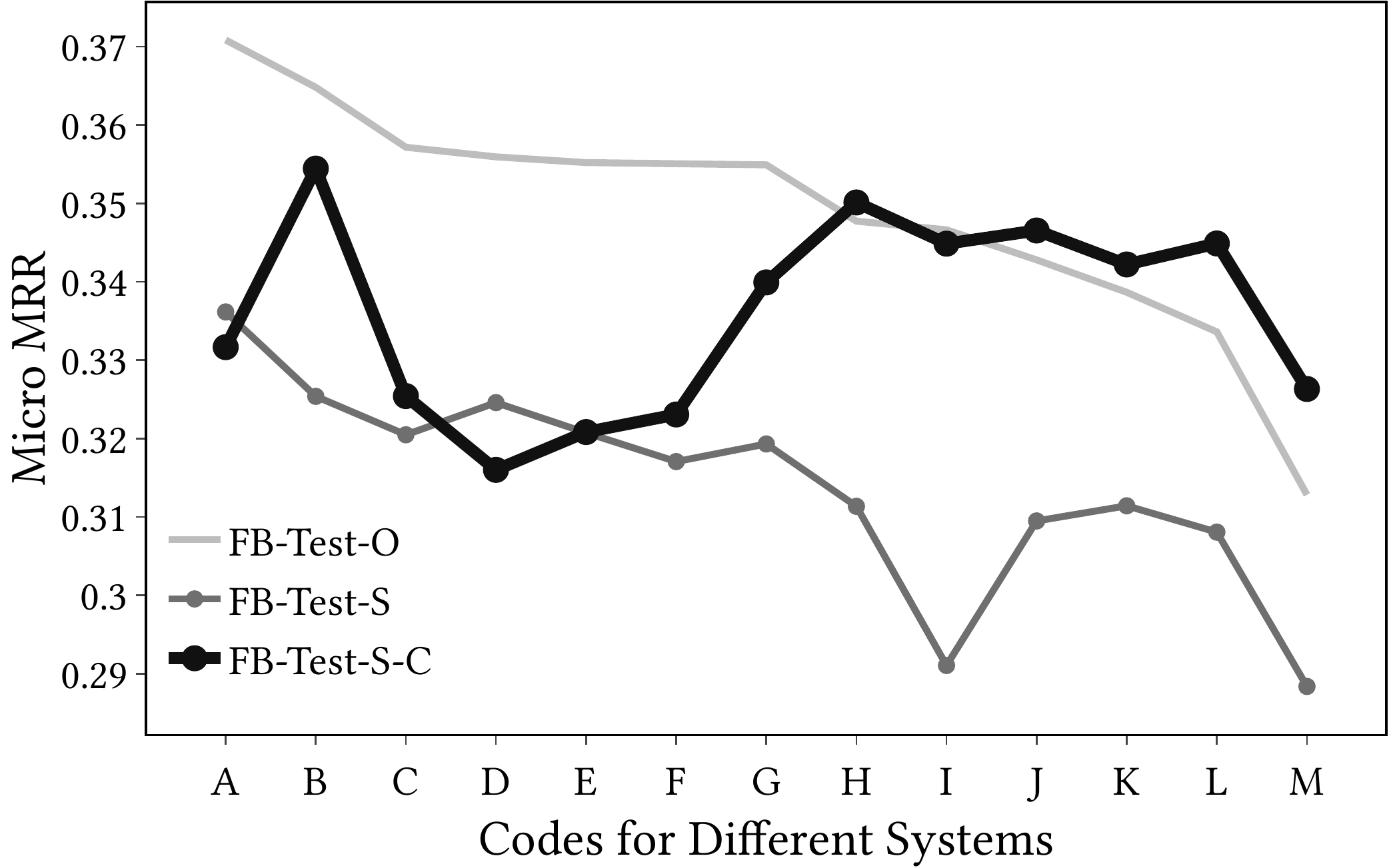}
\caption{Effect of label sparsity. Micro MRR values of different KGC models obtained on the original sparse labels (\texttt{FB-Test-O}) are in decreasing order. The $\tau$ correlation between \texttt{FB-Test-S} and the entire original test set \texttt{FB-Test-O} is 0.7949. The use of added dense labels (\texttt{FB-Test-S-C}) leads to obvious system ranking change with Kendall's $\tau$ of -0.2308 to \texttt{FB-Test-S}.}
\label{fig:rank_changes}
\end{figure}

\smallskip
\noindent
\textbf{Models.}
We implement 13 popular KGC models as follows:
\begin{itemize}[leftmargin=*]
\item {\bfseries Translational models} treat relation as the translation from source entity to target entity, including TransE~\cite{DBLP:conf/nips/BordesUGWY13}, RotatE~\cite{DBLP:conf/iclr/SunDNT19}, HAKE~\cite{DBLP:conf/aaai/ZhangCZW20}, and RotH~\cite{DBLP:conf/acl/ChamiWJSRR20}.

\item {\bfseries Decomposition models} match latent semantics of entity and relation to measure the plausibility, including DistMult~\cite{DBLP:journals/corr/YangYHGD14a}, RESCAL~\cite{DBLP:conf/icml/NickelTK11}, ComplEx~\cite{DBLP:conf/icml/TrouillonWRGB16}, TuckER~\cite{DBLP:conf/emnlp/BalazevicAH19}, and BiQUE~\cite{DBLP:conf/emnlp/GuoK21}.

\item {\bfseries Neural models} leverage neural networks to map entity and relation embeddings into matching scores, including ConvE~\cite{DBLP:conf/aaai/DettmersMS018}, CompGCN~\cite{DBLP:conf/iclr/VashishthSNT20}, InteractE~\cite{DBLP:conf/aaai/VashishthSNAT20}, and HittER~\cite{DBLP:conf/emnlp/ChenLG0ZJ21}.
\end{itemize}

Herein, for part of these KGC models, including TransE, RotatE, DistMult, ComplEx, RESCAL, and ConvE, we directly use the well-tuned checkpoints from LibKGE~\cite{DBLP:conf/iclr/RuffinelliBG20}, which can surpass the performance in the original papers. For the remaining seven KGC models, we reproduce their results using the released source code following the original papers.
For each KGC model, we repeatedly train for 10 times with the same hyper-parameters as in the original papers but with different random seeds, and then choose the one with the best micro MRR on the corresponding validation set.
According to our results, almost all models reach the performance reported in their original papers, and the maximum performance difference is only 0.002 in micro MRR.

\begin{figure}[t]
\centering
\includegraphics[width=\linewidth, page=1]{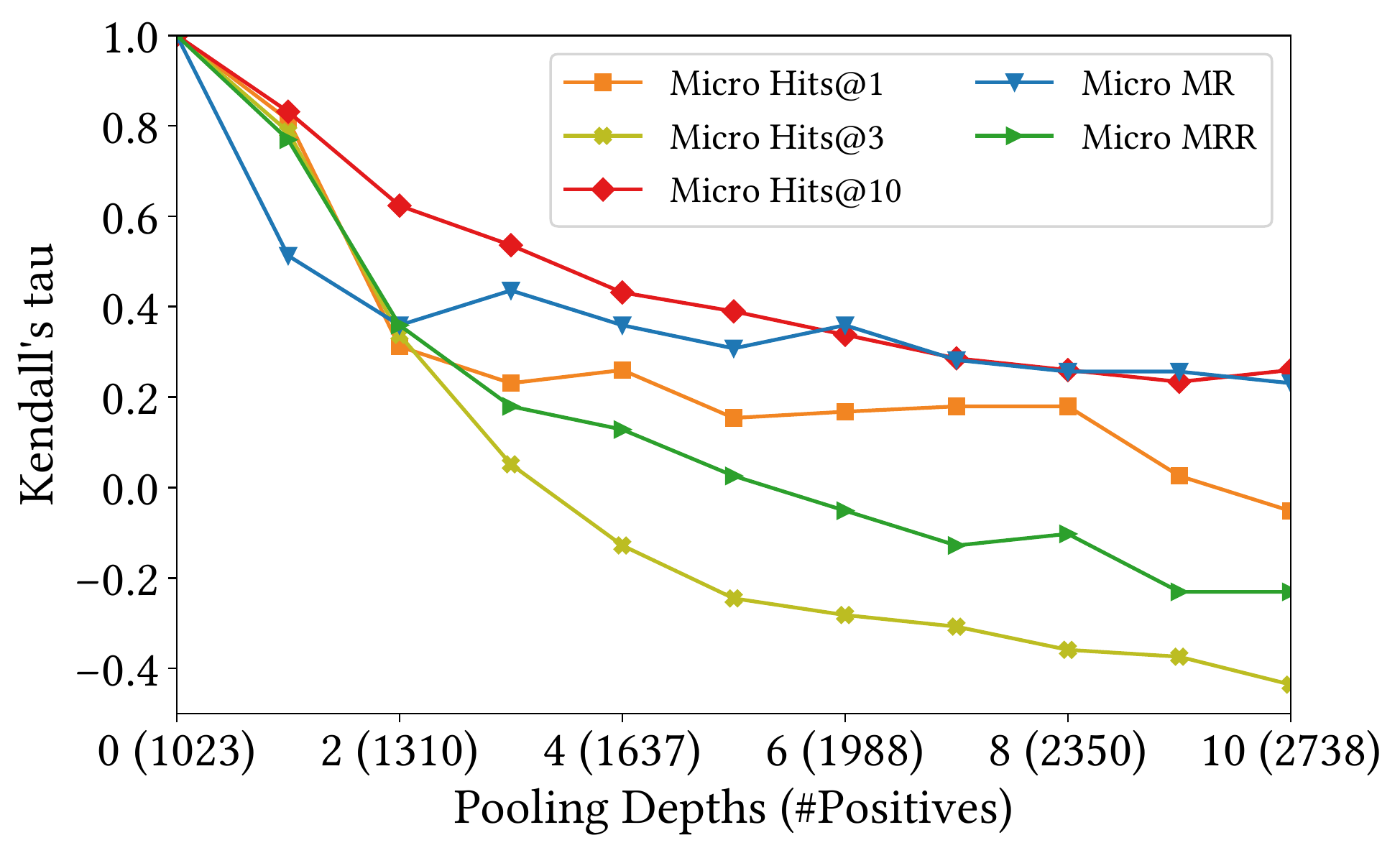}
\caption{Correlations of system rankings in micro metrics between \texttt{FB-Test-S} set and \texttt{FB-Test-S-C} with different pooling depths. The system ranking changes rapidly when adding even a few labeled triples (e.g., pool depth of 2). The correlation coefficient even decreases to negative in terms of Hits@1, Hits@3 and MRR with increasing pooling depth.}
\label{fig:pooling_depth}
\end{figure}

\subsection{Results} \label{sec.sparsity.results}
\textbf{The system ranking drastically changes with the increased label completeness.}
As mentioned, we include 13 popular KGC models to empirically examine whether label sparsity indeed affects the evaluation of KGC models.
The correlation between the system rankings obtained by the original \texttt{FB-Test-S} subset and by our annotated \texttt{FB-Test-S-C} with different pooling depths is summarized in Figure~\ref{fig:pooling_depth}.
It can be seen that, with the increase of label completeness (i.e., with larger pooling depth), the correlation of system rankings rapidly decreases for all micro metrics, which indicates a non-trivial change of system ranking brought by the use of more labels.
Specifically, in terms of \texttt{Micro MRR}, the Kendall's $\tau$ drops from 1.0 to 0.359 with merely 287 added labeled positive triples at pooling depth of 2.
Among these five micro metrics, in terms of \texttt{Micro Hits@1}, \texttt{Micro Hits@3}, and \texttt{Micro MRR}, the correlation of system rankings even drops to negative, while \texttt{Micro MR} and \texttt{Micro Hits@10} produce the most correlated results between ``complete'' and ``incomplete'' sets.
A likely cause is that \texttt{Micro MR} is defined as the rank of true entity, which is less sensitive to the top-ranked true answers but is more dominated by the bottom-ranked outliers.
And \texttt{Hits@K} shows to be largely affected by the choice of K, wherein the most correlated results come with $K=10$ and the worst correlation is produced with $K=3$.
Additional to correlation of system rankings, we present an example in Figure~\ref{fig:rank_changes}, wherein the Kendall's $\tau$ between the system rankings in \texttt{Micro MRR} obtained on the sparse and the pooled dense labels (depth=10) is only -0.2308. It can be seen that the system ranking dramatically changes when switching from the original sparse labels to our more complete labels. Models are coded in A-M since our focus is the evaluation protocol rather than benchmarking current methods. 
Notably, model A, the 1st on the \texttt{FB-Test-O} labels, is only ranked in the middle (the 8th) on the complete labels. Meanwhile, model D drops from the 4th to the 13th (the last), and model L moves from the 12th to the 4th.
When taking a closer look at the top-10 predictions for model L, we find that 12.5\% are predicted positives, in which more than half (6.7\%) are newly added annotations by the pooling.
By contrast, only 5.5\% out of 11.8\% of the predicted positives are newly labeled for model A (5.6\% out of 11.5\% for model D). From these analyses, it is obvious that false negatives have critical impact on the system rankings.
Indeed, these obvious changes of system ranking suggest the necessity of re-evaluating current KGC approaches on FB15K-237, the most commonly used dataset, with increased label completeness from TREC pooling.

\begin{table}[t]
\centering         
\caption{Correlations of system rankings by relation category between \texttt{FB-Test-S-C} and \texttt{FB-Test-S}. 
}
\begin{tabular}{ll|*{4}{c}}
\toprule
\multicolumn{2}{l|}{\multirow{2}{*}{\bfseries Categories}} &
\multicolumn{4}{c}{\textbf{Micro Average}} \\
& & {\bfseries MRR} & {\bfseries Hits@1} & {\bfseries Hits@3} & {\bfseries Hits@10} \\
\midrule
\multirow{4}{*}{\bfseries Head} & {\bfseries 1-1} & 1.0 & 1.0 & 1.0 & 1.0 \\
& {\bfseries 1-N} & 0.0256 & 0.0748 & 0.3731 & 0.4665 \\
& {\bfseries N-1} & 0.1026 & 0.0996 & 0.1782 & 0.2857 \\
& {\bfseries N-N} & -0.3333 & -0.0645 & -0.4698 & -0.1177\\
\midrule
\multirow{4}{*}{\bfseries Tail} & {\bfseries 1-1} & 1.0 & 1.0 & 1.0 & 1.0 \\
& {\bfseries 1-N} & -0.1026 & 0.1353 & 0.1988 & -0.1420 \\
& {\bfseries N-1} & 0.5641 & 0.7114 & 0.6123 & 0.7371 \\
& {\bfseries N-N} & -0.1282 & 0.0260 & -0.2746 & 0.3007 \\
\bottomrule
\end{tabular}
\label{tab:correlation_each_type}
\end{table}

\smallskip
\noindent{\bfseries Label sparsity does affect all relation categories, apart from one-to-one category.}
We further investigate the changes in system ranking by different relation categories when the label completeness is improved.
The relations in KG can be commonly categorized as one-to-one (1-1), one-to-many (1-N), many-to-one (N-1), and many-to-many (N-N) according to the average number of objects per subject and the average number of subjects per object~\cite{DBLP:conf/aaai/WangZFC14}.
Moreover, each question is either a Head question that predicts the subject $(?, p, o)$ or a Tail question that predicts the object $(s, p, ?)$.
As seen in Table~\ref{tab:correlation_each_type}, comparing the correlations of systems rankings from \texttt{FB-Test-S-C} to \texttt{FB-Test-S}, the system ranking is fixed on 1-1 category (correlation = 1), while changes drastically on N-1 (Head), 1-N (Tail) and N-N (Head and Tail) categories, where even negative correlation can be observed. 
Herein, the above results are due to the fact that the 1-1 category has no sparsity issue, while N-1 (Head), 1-N (Tail) and N-N (Head and Tail) categories had many true answers which are added by our annotations. Those added labels are indeed shown to have important influence on the evaluation. As a result, non-negligible changes of system ranking are observed.
However, the situations on 1-N (Head) and N-1 (Tail) categories are a little different. 
In principle, both should have little change in the correlation since there is only one answer as given by the original labels, but the correlations on 1-N (Head) in terms of \texttt{Micro MRR} and \texttt{Micro Hits@1} drastically decrease to close to zero.
As our annotations only add more true \textit{tail} entities for 1-N questions and true \textit{head} entities for N-1 questions, the difference on 1-N (Head) or N-1 (Tail) categories between \texttt{FB-Test-S-C} and \texttt{FB-Test-S} sets is only the scale (i.e., the number of associated labeled questions like $(?, p, o')$ where $o'$ are newly labeled) of the triple questions.
Thereby, the changes on 1-N (Head) and N-1 (Tail) categories could attribute to that {\itshape micro} metrics may not stable on the size of test set, which is further analyzed in Section~\ref{sec.scale.stability}.

\begin{figure}[t]
\centering
\includegraphics[width=\linewidth, page=1]{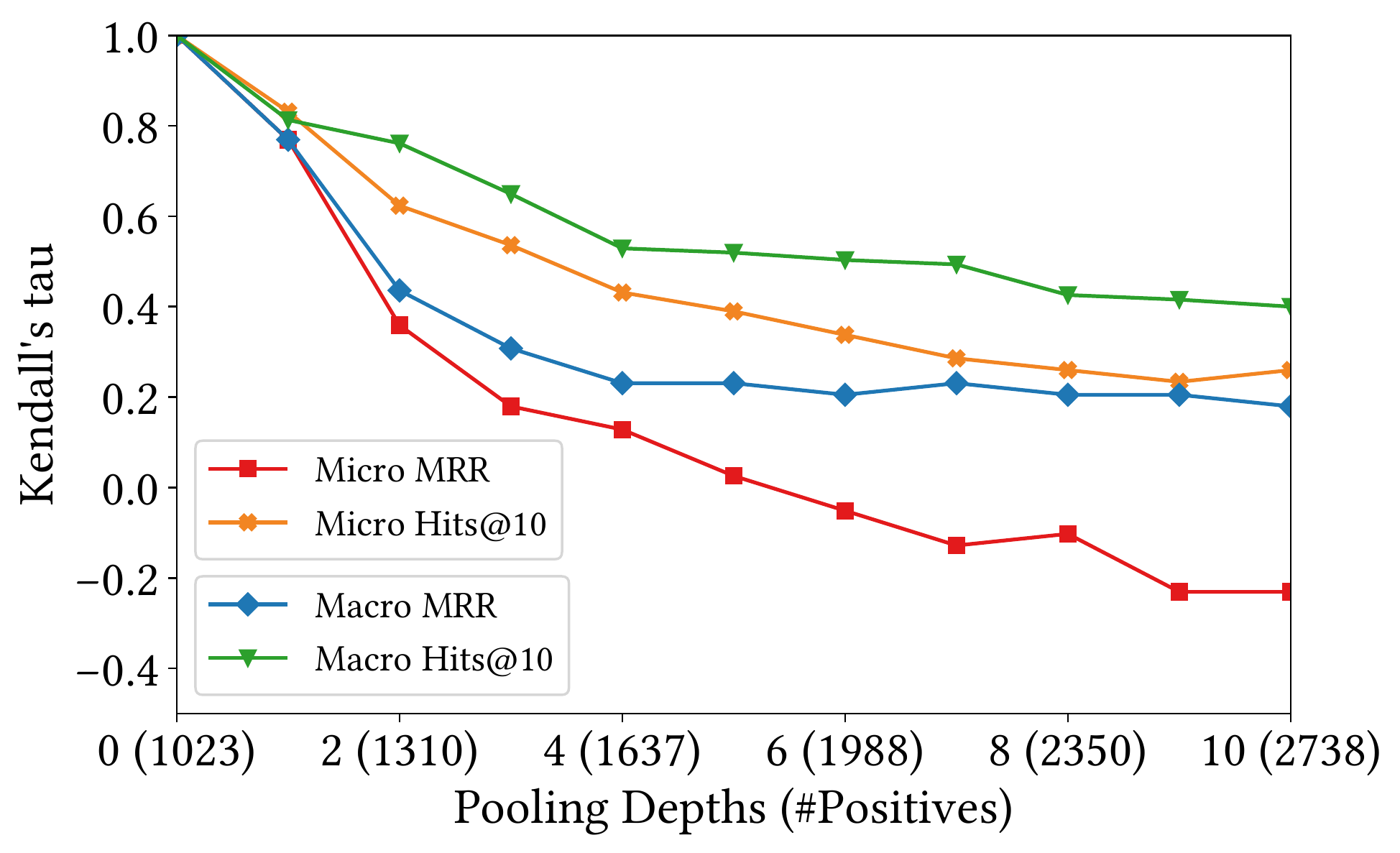}
\caption{Correlations of system rankings between \texttt{FB-Test-S} and \texttt{FB-Test-S-C} with different pool depths using both \textit{micro} and \textit{macro} metrics. Comparing to micro metrics, the system rankings by macro metrics are more stable.}
\label{fig:pooling_depth_macro}
\end{figure}

\smallskip
\noindent\textbf{Macro metrics are more stable to label sparsity.} \label{sec.influence_to_macro}
With the adaption of IR-like evaluation, we can introduce macro metrics to the KGC task.
To assess the influence of label sparsity on macro metrics, we compare the changes in system ranking between the micro and macro settings.
Therefore, we choose \texttt{MRR} and \texttt{Hits@10} metrics to study the differences under micro or macro settings, as both are widely-used in previous works.
The change curve of the correlation of system rankings under \texttt{MRR} and \texttt{Hits@10} metrics on \texttt{FB-Test-S-C} set with different pooling depths is shown in Figure~\ref{fig:pooling_depth_macro}.
We can see that the correlations on all metrics decrease when the pooling depth increases, while macro metrics (namely, \texttt{Macro MRR} and \texttt{Macro Hits@10}) can achieve higher correlations than micro metrics (namely, \texttt{Micro MR} and \texttt{Micro Hits@10}) at any pooling depth. For example, when the pooling depth is 10, the gap between correlations of \texttt{Micro MRR} and \texttt{Macro MRR} is 0.41, and the gap for \texttt{Hits@10} is 0.14.
These results are consistent with our previous observations in Section~\ref{sec.sparsity.results}, namely, label sparsity indeed causes changes in the system ranking.
But meanwhile, we can also see that macro metrics show the ability in negating the effect of label sparsity in that macro metrics produce relatively correlated system rankings between ``complete''  and sparse labels.
Note that evaluation with micro metrics is still desirable since macro metrics may overemphasize one-to-one relations. Therefore, our suggestion is to report both \textit{micro} and \textit{macro} metrics to reflect both the per-answer and per-question performance.

\begin{table}[t]
\centering         
\caption{Correlations between system rankings obtained on \texttt{FB-Test-S} and \texttt{FB-Test-S-C}. On \texttt{FB-Test-S-C} set, micro metrics are computed with different relation ratios derived from the relation distribution on \texttt{FB-Test-S-C} set (Complete) or \texttt{FB-Test-S} set (Sparse).}
\resizebox{0.99\linewidth}{!} {%
\begin{tabular}{l|*{5}{c}}
\toprule
{\bfseries Distribution} & {\bfseries MR} & {\bfseries MRR} & {\bfseries Hits@1} & {\bfseries Hits@3} & {\bfseries Hits@10} \\
\midrule
Complete & 0.2308 & -0.2308 & -0.0519 & -0.4358 & 0.2598 \\
Sparse & 0.2308 & -0.2051 & 0.0256 & -0.2051 & 0.3377 \\
\bottomrule
\end{tabular}
}
\label{tab:changing_distribution}
\end{table}

\smallskip
\noindent
{\bfseries Changes in relation distribution is not the main cause for the changes in system rankings.}
The added labels increase the proportion of the one-to-many, many-to-one, and many-to-many relations in the test set.
As shown in Figure~\ref{fig:relation_distribution}, the relation distribution of \texttt{FB-Test-S-C} slightly changes relative to that of \texttt{FB-Test-S} after complementing the judgment of missing triples, which raises the concern about whether the altered relation distribution is the main reason of the drastic changes of system ranking.
Thereby, to answer this question, we examine the effect of relation distribution by using the relation ratio according to the \texttt{FB-Test-S} set when computing the final metrics.
In other words, we try to maintain the ratios of relations in the final evaluation on \texttt{FB-Test-S-C}, namely, changing the $\frac{c_p}{|\mathcal{K}_{test}|}$ item in Eq.~\ref{eq.micro} to the corresponding occurrences ratios in the original \texttt{FB-Test-S} set.
As seen in Table~\ref{tab:changing_distribution}, although using original ``sparse'' distribution to compute the metrics, the system ranking still changes a lot, namely, the correlation coefficients for all micro metrics are still much less than 0.4.
Accordingly, the results suggest that changes in relation distribution in the \texttt{FB-Test-S-C} test set do not contribute much to the changes in the system rankings.

\begin{figure}[t]
\centering
\includegraphics[width=\linewidth, page=1]{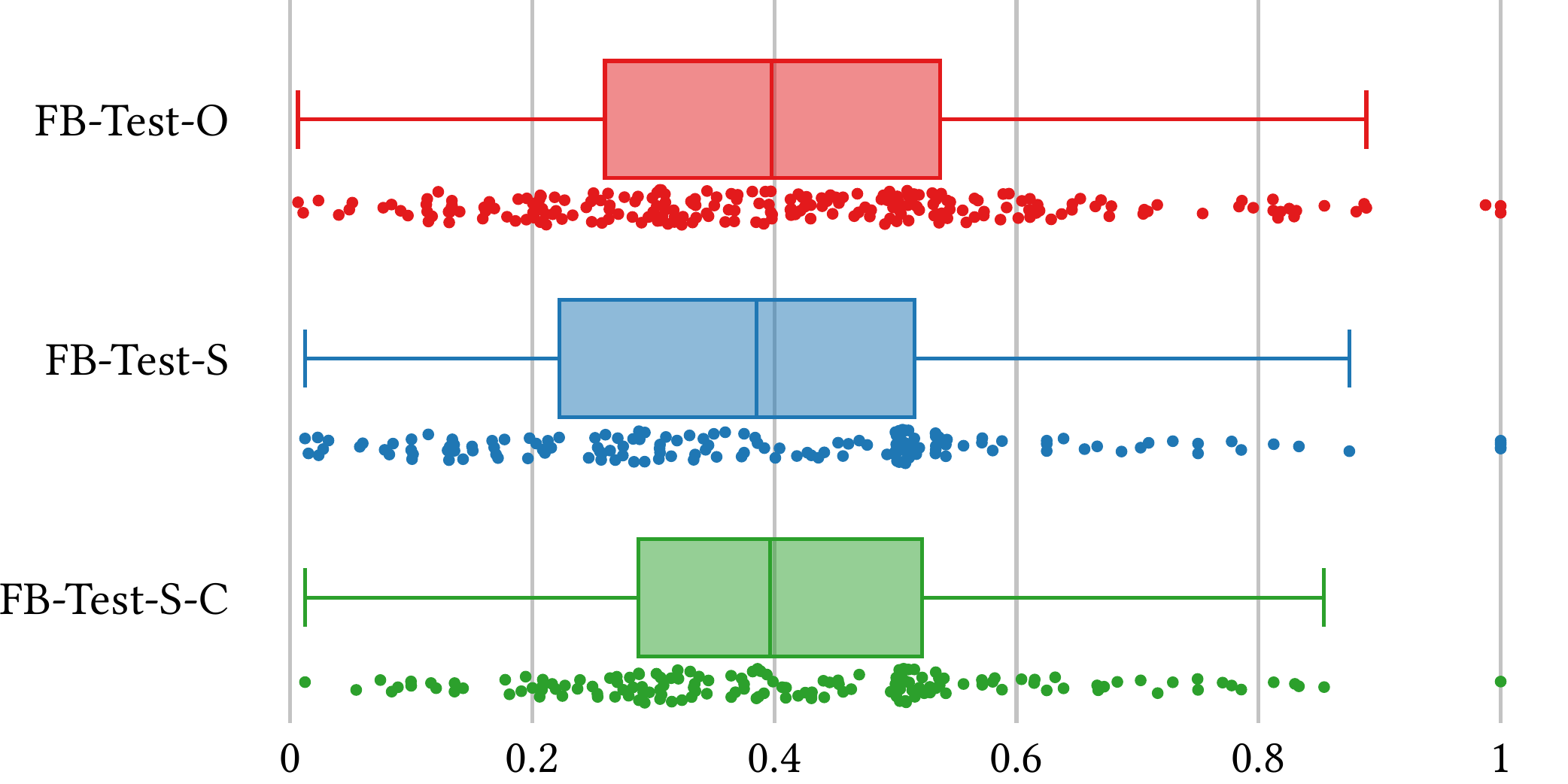}
\caption{Performance of HittER~\cite{DBLP:conf/emnlp/ChenLG0ZJ21} on different relations in Micro MRR. The median Micro MRR values on \texttt{FB-Test-O} and \texttt{FB-Test-S-C} sets are similar while is slightly smaller on \texttt{FB-Test-S} set.}
\label{fig:difficulty}
\end{figure}

\smallskip
\noindent
{\bfseries Our supplemental annotated triples are not easy-to-predict.}
Previous work~\cite{DBLP:conf/emnlp/SafaviK20} has indicated that a reasonable test set should not be easy-to-predict.
Herein, to verify the difficulty of our annotated triples, we use the state-of-the-art HittER model~\cite{DBLP:conf/emnlp/ChenLG0ZJ21}, to rank entities grouped by relations.
Figure~\ref{fig:difficulty} shows the performance of HittER on different relations in terms of \texttt{Micro MRR}.
We can see that the median micro MRR values on the \texttt{FB-Test-O} and \texttt{FB-Test-S-C} sets are much similar, while a slightly smaller median is produced on the sampled \texttt{FB-Test-S} set which may come from the sampling bias.
Thereby, these three test sets are more or less of the same level of difficulty.
In other words, our annotated triples are not easy-to-predict for a state-of-the-art KGC model, and the evaluation on our relatively ``complete'' \texttt{FB-Test-S-C} set is reasonable.

\begin{figure*}[t!]
\centering
\begin{subfigure}{.31\textwidth}
\includegraphics[width=1.0\linewidth, page=1]{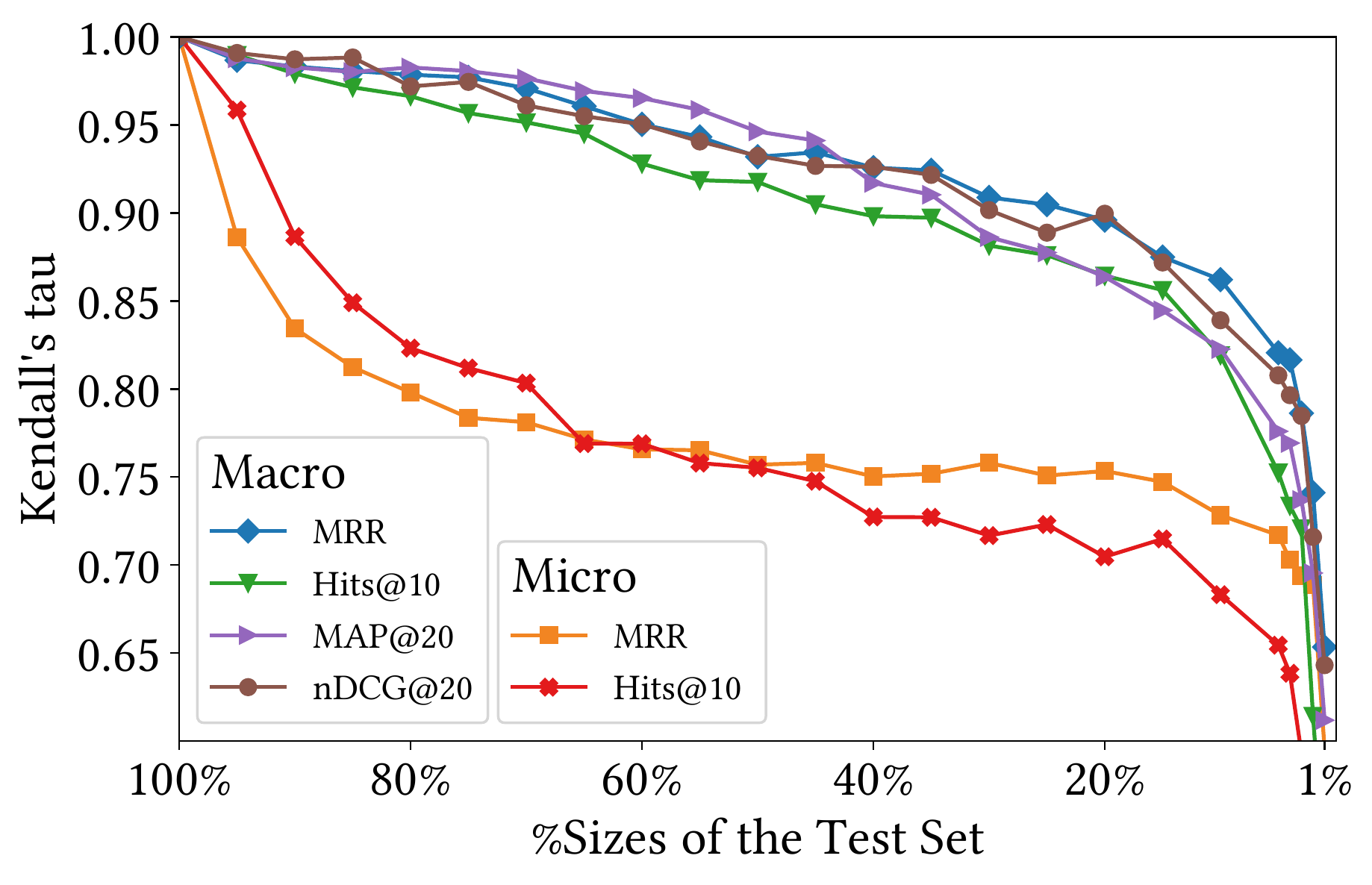}
\caption{FB15k-237 (\texttt{FB-Test-O})}
\label{fig:sub_stability_FB15k}
\end{subfigure}
\begin{subfigure}{.31\textwidth}
\includegraphics[width=1.0\linewidth, page=2]{figures/stability.pdf}
\caption{WN18RR}
\label{fig:sub_stability_WN18RR}
\end{subfigure}
\begin{subfigure}{.31\textwidth}
\includegraphics[width=1.0\linewidth, page=3]{figures/stability.pdf}
\caption{\texttt{FB-Test-S-C}}
\label{fig:sub_stability_complete}
\end{subfigure}
\caption{Correlations of system rankings on different sizes of test set relative to the whole original test set. Higher Kendall's $\tau$ indicates better stability to the change of test data.}
\label{fig:stability}
\end{figure*}

\smallskip
\noindent{\bfseries Answer to RQ1.}
Based on the observation on \texttt{FB-Test-S-C} set, we can conclude that label sparsity in KGC test set does affect the evaluation results of entity ranking regardless of micro or macro metrics are adopted. Even though, the macro metrics are shown to be more robust. In the next section, we further compare the use of macro and micro metrics to examine how they behave similarly or differ when used for KGC evaluation.

\section{Study on Evaluation Metric} \label{sec.metric.study}
The preceding Section~\ref{sec.sparsity.results} has shown that macro metrics are more stable to label sparsity than micro metrics. Since KGC is essentially a ranking task with question answering nature \cite{DBLP:conf/rep4nlp/WangRGBM19}, question-wise macro metrics measure how well a KGC model answers a given question, unlike the micro metrics that measure how a specific answer is identified while ignoring other valid answers. 
As suggested by \citet{DBLP:conf/sigir/Sakai06,DBLP:conf/acl/Sakai20}, when choosing a metric for evaluation, it is expected to be stable and not sensitive to test data, namely, producing as consistent results as possible on different test data without losing the ability in discriminating different systems. Therefore, to answer RQ2 that how do macro metrics differ from micro metrics, especially that whether the introduction of the per-question macro metrics could be beneficial, we compare their stability and discriminative power in this section.

\subsection{Experimental Design}
Similar to the study on label sparsity in the previous section, analyses are carried out on the same 13 KGC models as described in Section~\ref{sec.sparsity.design}. 
In addition to the original FB15k-237 test set and our annotated \texttt{FB-Test-S-C} set, we also use the popular WN18RR~\cite{DBLP:conf/aaai/DettmersMS018} dataset for the study on metrics in this section.
As for the evaluation metrics, additional to \texttt{Micro Hits@10}, \texttt{Micro MRR}, \texttt{Macro Hits@10} and \texttt{Macro MRR}, we also borrow two other macro metrics widely used in IR evaluation, \texttt{MAP@20}~\cite{DBLP:books/daglib/0021593}
and \texttt{nDCG@20}~\cite{DBLP:journals/tois/JarvelinK02}.
It should be noted that \texttt{bpref}~\cite{DBLP:conf/sigir/BuckleyV04} is not studied because the original FB15k-237 and WN18RR datasets do not come with any labeled negative triple.

\smallskip
\noindent
\textbf{Data Sensitivity.}
Similar to studies in IR evaluation \cite{DBLP:conf/sigir/BuckleyV04,DBLP:conf/sigir/ButtcherCYS07,DBLP:conf/sigir/HeMO08,DBLP:conf/ictir/RahmanKEL20}, we simulate a series of test subsets in different sizes (1\%-95\% of the total triples) by randomly discarding a proportion of triples from the test set. 
Then, we analyze the system rankings on each test subset using various metrics, and study how the system rankings change on these test subsets in different sizes, by computing the Kendall's $\tau$ between system rankings on test subsets and the whole test set.
Besides, to eliminate the sampling bias, we repeat random discarding for 50 times on each sampling size, and report the average results as the final correlation.

\smallskip
\noindent \textbf{Discriminative Power.}
When comparing the performance of two systems, statistical significance test is usually conducted.
A significant difference (i.e., a $p$-value smaller than a given threshold) between two systems confirms that the better performance of one system over the other is not obtained by chance.
Akin to \citet{DBLP:conf/sigir/Sakai06}, given a set of systems and a ranking metric, the discriminative power of this metric is defined as $p$-values on every system pair using a pairwise significance test. Herein, the paired two-tailed Student t-test is used in this work.
Highly discriminative metrics are those who can obtain as many small $p$-values as possible~\cite{DBLP:conf/acl/Sakai20,DBLP:conf/ecir/Sakai21}.
It should be noted that a highly discriminative metric does not always lead to correct conclusions since statistical significance is not necessarily correct~\cite{DBLP:conf/acl/Sakai20}, but we can still consider these metrics to be more sensitive to changes in model performance. Combined with the analysis on data sensitivity, we expect the results to give a clue about what evaluation paradigms and metrics are suitable for future KGC research.

\begin{figure*}[t!]
\centering
\begin{subfigure}{.31\textwidth}
\includegraphics[width=1.0\linewidth, page=1]{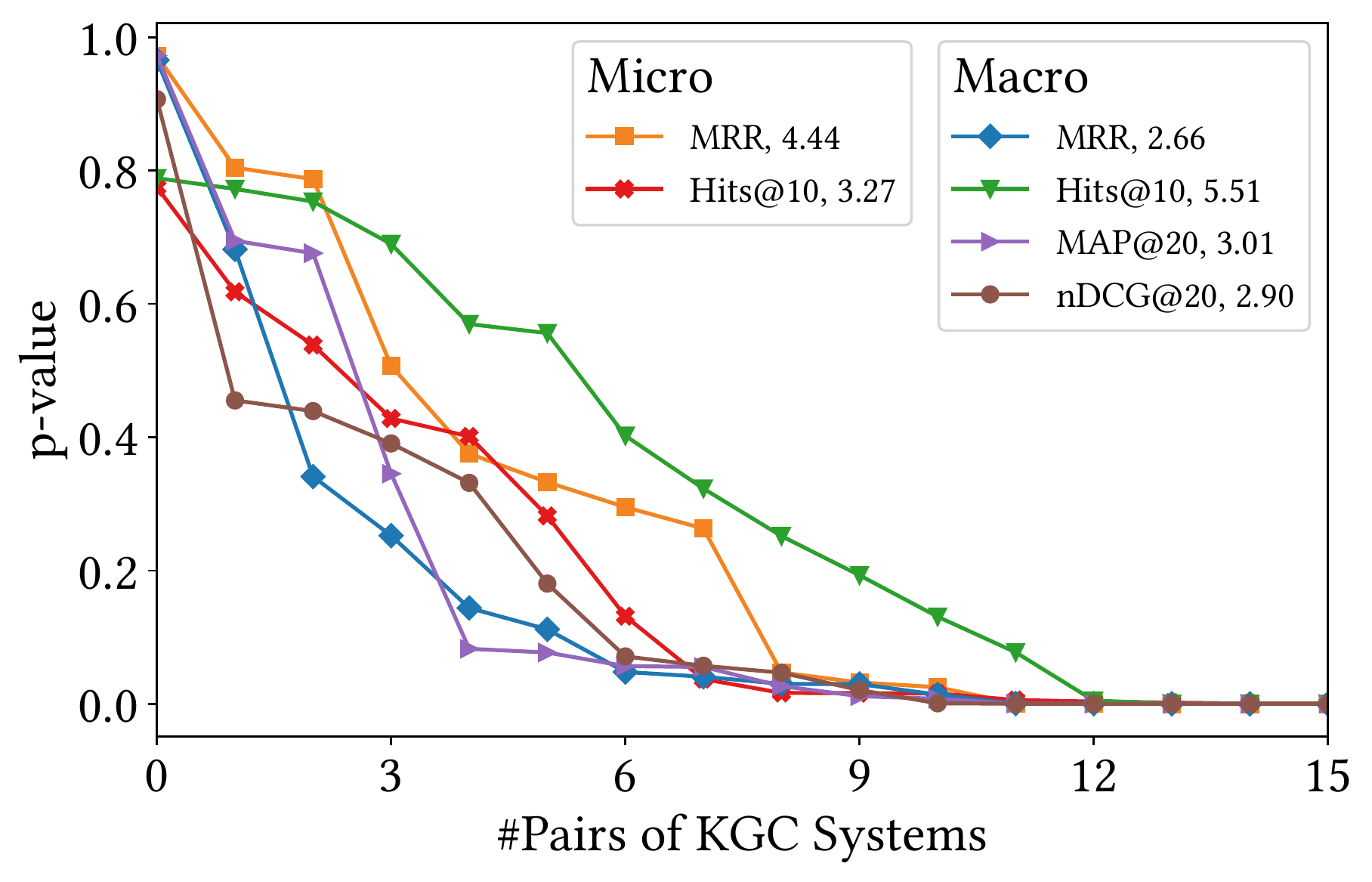}
\caption{FB15k-237 (\texttt{FB-Test-O})}
\label{fig:sub_discriminative_origin}
\end{subfigure}
\begin{subfigure}{.31\textwidth}
\includegraphics[width=1.0\linewidth, page=2]{figures/discriminative_power.pdf}
\caption{WN18RR}
\label{fig:sub_discriminative_wn}
\end{subfigure}
\begin{subfigure}{.31\textwidth}
\includegraphics[width=1.0\linewidth, page=3]{figures/discriminative_power.pdf}
\caption{\texttt{FB-Test-S-C}}
\label{fig:sub_discriminative_complete}
\end{subfigure}
\caption{Discriminative power analysis. The p-values between all pairs of KGC models are in descending order. Smaller area under the curve, as given next to the metric in the legend, indicates the relatively stronger discriminative power.}
\label{fig:discriminative_power}
\end{figure*}

\subsection{Results}\label{sec.scale.stability}
\textbf{Macro metrics are more stable than micro metrics on incomplete test graph.}
The correlations of system rankings under different sample sizes are summarized in Figure~\ref{fig:stability}.
On FB15k-237 test set, as seen in Figure~\ref{fig:sub_stability_FB15k}, all macro metrics (including \texttt{Macro MRR}, \texttt{Macro Hits@10}, \texttt{nDCG@20}, and \texttt{MAP@20}) show strong stability across different sample sizes of test set, while micro metrics like \texttt{Micro MRR} and \texttt{Micro Hits@10} produce very different system rankings on the FB15k-237 test set when the sample size decreases.
Specifically, when the size of the sampled test set decreases to 80\%, Kendall's $\tau$ values of \texttt{Micro MRR} and \texttt{Micro Hits@10} are about 0.8, while almost all the $\tau$ values of the macro metrics are above 0.95.
Then, as seen in Figure~\ref{fig:sub_stability_WN18RR}, the stability results on WN18RR test set are a little different. We can see that micro metrics are relatively stable on different sizes of test set, likely due to the more ``complete'' nature of the domain specific knowledge graph WordNet, and containing more one-to-one relations in the test set. 
Even though, on WN18RR test set, slightly but consistently better stability of the macro metrics are still observed. 
Finally, according to Figure~\ref{fig:sub_stability_complete}, on our ``complete'' \texttt{FB-Test-S-C} set, it can be seen that the micro metrics are almost as stable as macro ones, while the \texttt{Macro MRR} is still relatively more stable, namely, the curve of \texttt{Macro MRR} is above its micro counterpart.
Note that among all metrics on \texttt{FB-Test-S-C}, \texttt{Micro Hits@10} shows to be the most stable, suggesting that TREC-style KG augmentation indeed improves the evaluation stability of micro metrics.
These above observations indicate that the stability of evaluation metric is related to the completeness of a KG test set, especially for the micro metrics. Despite the better stability macro metrics exhibit, as mentioned, it is necessary to take also the discriminative power of different metrics into consideration.

\smallskip
\noindent
\textbf{Both micro and macro metrics are discriminative, but macro metrics are better on incomplete test graph.}
The curves of p-values of all system pairs in a descending order on FB15k-237, WN8RR and \texttt{FB-Test-S-C} are summarized in Figure~\ref{fig:discriminative_power}, wherein the smaller area under the curve of a metric indicates the relatively higher discriminative power of this metric.
As shown in Figure~\ref{fig:discriminative_power}, the main finding is that almost all macro metrics are more discriminative than micro metrics on incomplete test set (like FB15k-237), but the gap between micro and macro metrics decreases when the test set becomes ``complete'' (like \texttt{FB-Test-S-C}).
As for \texttt{Hits@10}, we observe that it shows inconsistent changes in discriminative power between macro and micro settings compared with other metrics like \texttt{MRR}.
We find that some triple questions from test set (especially on WN18RR) come with relatively bottom ranks for true entities, namely, the mean rank of all 13 models on WN18RR is around 3000 and the value on FB15k-237 and \texttt{FB-Test-S-C} is about 200.
Thus, the micro and macro Hits@10 may not be sensitive to bottom ranks since all answers with ranking larger than 10 are merely ignored.
The same conclusion can also be drawn from Figure~\ref{fig:sub_stability_WN18RR}, wherein both \texttt{Micro Hits@10} and \texttt{Macro Hits@10} show to be less stable than other metrics.
Overall, results show that both micro and macro metrics are discriminative on a ``complete'' KGC test set, but micro metrics may less be capable of significantly differentiating the performance of KGC models on an incomplete test set.

\smallskip
\noindent{\bfseries Answers to RQ2\&3.}
Based on the above analyses, for RQ2, we can see that the current micro metrics differ from macro metrics in that: (1) more affected by label sparsity; (2) more sensitive to change of test data; (3) not as sensitive to model performance as the macro metrics.
Therefore, for RQ3, we suggest to employ TREC-style pooling to negate the impact of label sparsity. In addition to the widely used \textit{micro} metrics, we recommend to report also the \textit{macro} metrics to reflect both per-answer and per-question KGC performance. Moreover, on datasets with relatively complete labels (e.g., our labeled pool), or with dominating one-to-one relations (e.g., WN18RR), since micro metrics behave similarly to macro metrics, we recommend to report only \textit{macro} metrics for better stability, discriminative ability, and less sensitivity to label sparsity.

\section{Related Work}\label{sec.related.work}
\subsection{Problems behind Entity Ranking}
Recently, quite a few works have raised the hidden problems behind entity ranking protocol.
\citet{DBLP:conf/cikm/AkramiGHL18,DBLP:conf/sigmod/AkramiSZHL20} conduct studies to analyze the influence of unrealistic triple whose the relation is an anomaly (near-reverse, near-same, or Cartesian product), and find that the current evaluations are much less compatible to reflect the model performance, and demonstrate that a re-investigation of possible effective approaches other than current entity ranking are indeed needed for following KGC researches.
\citet{DBLP:conf/acl/SunVSTY20} argue that the way to select ranking position of a triple which has ties can drastically affect the model performance.
Their main conclusion is that current entity ranking protocol is problematic in terms of data distribution and metric calculation, but they do not consider the label sparsity phenomenon.
Meanwhile, recent studies~\cite{DBLP:conf/akbc/PezeshkpourT020,DBLP:conf/akbc/SperanskayaS020,DBLP:conf/rep4nlp/WangRGBM19} have revealed that the performance of KGC models may be misled by unobserved true (positive) triples, which are always treated as negatives in the current entity ranking protocol.
Thereby, in spite of the advancements in effectiveness~\cite{DBLP:conf/acl/ChamiWJSRR20,DBLP:conf/emnlp/ChenLG0ZJ21}, it also needed to investigate whether and how these false negative triples affect the evaluation of KGC systems, and whether the entity ranking protocol on the current ``incomplete'' test set corresponds to the actual prediction capability of KGC models.
To the best of our knowledge, this paper is the first to systematically investigate the label sparsity issue in KGC evaluation.

\subsection{Alternative Evaluation Protocols}
Apart from the entity ranking protocol, other evaluation protocols have been proposed for the KGC task.
A few works~\cite{DBLP:conf/nips/BordesUGWY13,DBLP:conf/aaai/WangZFC14,DBLP:conf/aaai/WangGL18} experiment with the triple classification protocol and find it easy to get a high classification accuracy on sampled negative triples.
However, \citet{DBLP:conf/emnlp/SafaviK20} gather a novel KGC dataset with human-labeled hard negatives, suggesting that triple classification is far from being solved.
Moreover, \citet{DBLP:conf/akbc/PezeshkpourT020} redefine the triple classification task and introduce YAGO3-TC dataset with a similar TREC-style annotation method, but only 2 baselines were used to pool triples.
Besides, \citet{DBLP:conf/rep4nlp/WangRGBM19} propose Pair-entity Ranking (PR) wherein a weighted relation-averaged MAP on every single predicate $p$ is used, and the weight of $p$ is based on the total number of the occurrences in the test set, namely, predicting all possible triples given (?, $p$, ?).
\citet{DBLP:conf/sigir/MeiZMD18} present Max-K criterion for entity ranking that ask KGC models to give at most K candidates as the predicted answers for each question, wherein macro averaged recall, precision, and F1 are calculated to measure the model performance.
Notably, our work shares common views on the definition of macro metrics with~\cite{DBLP:conf/sigir/MeiZMD18}, but differs in that \citet{DBLP:conf/sigir/MeiZMD18} apply macro metrics in a Max-K setting, while our work mainly compares the use of macro and micro metrics for entity ranking evaluation over label sparsity, data sensitivity, and discriminative power.
Meanwhile, other works focus on refining or extending the entity ranking evaluation by proposing new evaluation metrics.
\citet{DBLP:conf/www/TiwariBR21} present the weighted geometric mean rank (WMR) that considers the difficulty of ranking for each relation, and argue that anomaly, test data distributions, and lacking of statistical significant test in entity ranking can hinder the understanding of the accuracy of current models.
\citet{berrendorfAmbiguity2020} propose the adjusted mean rank (AMR) that adjusts the MR for chance.
The main difference between these work with ours is that we aim to investigate label sparsity in the current entity ranking protocol, rather than proposing a new evaluation protocol.

\subsection{Judgement Incompleteness in IR Evaluation}
As manually labeling all query-document pairs is often infeasible, IR evaluation has long been affected by the label sparsity problem. Apart from web search engines that infer relevance from huge amount of query logs, e.g., by the click rate per impression~\cite{DBLP:books/daglib/0021593}, IR research has advanced from the Cranfield paradigm~\cite{DBLP:conf/sigir/Cleverdon91} to adopt the TREC pooling method~\cite{voorheesTREC2005} to balance between label completeness and manpower cost~\cite{DBLP:journals/corr/abs-2109-00062,DBLP:conf/ictir/RahmanKEL20}.
Specifically, for a given set of test queries that requires human annotations, the top-\textit{k} documents returned by a diverse set of IR systems are merged into the so-called TREC pool, where \textit{k} is the pooling depth. Relevance assessments are then conducted on this pool to produce the final human labels.
Although TREC pools are still incomplete given the likely existence of many unlabelled positive documents in large-scale collections, they have been shown to be reliable in comparing IR systems using early precision metrics like (macro) MAP@20. The TREC pools have been used as the ``complete'' labels sets in many studies on IR evaluation, especially on the label sparsity~\cite{DBLP:conf/sigir/HeMO08,DBLP:conf/sigir/ButtcherCYS07,DBLP:conf/sigir/BuckleyV04,DBLP:conf/www/Kirnap0BECY21,DBLP:conf/ictir/RahmanKEL20,DBLP:journals/corr/abs-2109-00062}.

\section{Conclusions}\label{sec.conclusion}
In this paper, we systematically study whether and how the label sparsity affects the KGC evaluation. We first construct a relatively ``complete'' KGC test set by manual annotations following the TREC pooling method.
According to our analysis using 13 popular KGC models, the system ranking drastically changes with the increase of label completeness, demonstrating a strong impact of label sparsity.
Then, we further introduce IR-like macro-average metrics into the evaluation of KGC models, and experiments demonstrate that macro metrics are less sensitive to label sparsity and more stable and discriminative under different settings. Thus, we recommend reporting both micro and macro metrics to reflect the different aspects of the KGC task, while employing TREC-style pooling to deal with the label incompleteness. 
As label sparsity exists in quite a few ranking-oriented tasks such as multi-label classification~\cite{DBLP:conf/aaai/WuLG16,DBLP:journals/pr/LiWGZYJ16}, label distribution learning~\cite{DBLP:conf/ijcai/XuZ17}, and entity linking~\cite{DBLP:conf/emnlp/Rosales-MendezH19},
our findings may also suggest the need for TREC pooling in a wider range of research in order to create relatively complete datasets with limited human efforts.
In future work, we plan to extend our study to other IR metrics like bpref~\cite{DBLP:conf/sigir/BuckleyV04} and infAP~\cite{DBLP:conf/cikm/YilmazA06} for KGC evaluation. 
Other factors over meta-evaluation on metrics will also be examined, such as pairwise discriminative power~\cite{DBLP:conf/cikm/ChuMZ0SZM21}, user satisfaction~\cite{DBLP:conf/sigir/ZhangMLXMZM20,DBLP:conf/sigir/McDuff0CRC21,DBLP:conf/sigir/ChenZL0M17}, and diversity~\cite{DBLP:conf/sigir/SakaiZ19,DBLP:conf/sigir/AmigoSA18}.

\begin{acks}
We thank reviewers for their valuable comments and suggestions. We also thank Yujie Nie, Bo Li, Zhibo Hu, Linwei Che, and Xinlin Peng for their help during the data annotation procedure. This work is supported by the National Key Research and Development Program of China (Grants No. 2020AAA0106400), the National Natural Science Foundation of China (Grants No. U1936207), and the University of Chinese Academy of Sciences.
\end{acks}

\bibliographystyle{ACM-Reference-Format}
\balance
\bibliography{references}

\end{document}